\documentclass[runningheads,a4paper]{llncs}

\usepackage{amssymb}
\usepackage{amsmath}
\usepackage{graphicx}
\usepackage{epsfig}

\usepackage{url}
\urldef{\mailsa}\path|{klakhman, burtsev.m}@gmail.com|    
\newcommand{\keywords}[1]{\par\addvspace\baselineskip
\noindent\keywordname\enspace\ignorespaces#1}

\begin{document}
\mainmatter  

\title{Neuroevolution Results in Emergence of Short-Term Memory for \\ Goal-Directed Behavior\thanks{Manuscript was submitted to the 12th International Conference on the Simulation of Adaptive Behavior, 27--31 August 2012, Odense, Denmark}}
\titlerunning{Neuroevolution Results in Emergence of Short-Term Memory}

%
%
\author{Konstantin Lakhman\and Mikhail Burtsev}
\authorrunning{Konstantin Lakhman and Mikhail Burtsev}

\institute{Neuromorphic Cognitive Systems Lab,\\
Department of Neuroscience, Kurchatov NBIC-Centre,\\
National Research Center ``Kurchatov Institute'', Moscow, Russia\\
\mailsa\\}

%
%

\maketitle

\begin{abstract}
Animals behave adaptively in the environment with multiply competing goals. Understanding of the mechanisms underlying such goal-directed behavior remains a challenge for neuroscience as well for adaptive system research. To address this problem we developed an evolutionary model of adaptive behavior in the multigoal stochastic environment. Proposed neuroevolutionary algorithm is based on neuron's duplication as a basic mechanism of agent's recurrent neural network development. Results of simulation demonstrate that in the course of evolution agents acquire the ability to store the short-term memory and, therefore, use it in behavioral strategies with alternative actions. We found that evolution discovered two mechanisms for short-term memory. The first mechanism is integration of sensory signals and ongoing internal neural activity, resulting in emergence of cell groups specialized on alternative actions. And the second mechanism is slow neurodynamical processes that makes possible to code the previous behavioral choice.
\end{abstract}

\section{Introduction}

Understanding the mechanisms of learning and maintenance of effective behavior in stochastic environments with complex hierarchy of goals is one of key issues in the study of neural processing and for the development of bio-inspired artificial intelligence. However, these mechanisms are better understood if history of their development is taken into account.

Fields of machine learning and adaptive behavior have been rapidly developing in the recent decades. One of the most popular approaches in these areas is reinforcement learning (RL) \cite{SuttonBarto} and its modifications. It can be effectively used for learning of autonomous agents in many domains. However, problem spaces with multiple goals require execution of alternative behavioral strategies that are not possible for RL algorithms, since the behavior of the agent is determined by the single value function. Also, it is not always possible to define appropriate value function \textit{a priori}. To address this issue approaches for the generation of reinforcement values using evolutionary algorithms are currently under development \cite{SinghBarto}.

Other algorithms are also unable to produce effective solution for the environments with hierarchy of goals \cite{BotvinickBarto}. The various methods designed for the generation of behavioral sequences \cite{SandamirskayaSchoner,KomarovBurtsev} in most cases cannot effectively work in situations with a large number of goals present in the environment and when an agent has to autonomously explore them. Neuroevolutionary approaches have been successfully used for the synthesis of autonomous agent's controllers in low-formalized tasks \cite{FloreanoMondana,FloreanoMattiussi}.

To solve the problem of alternative behaviors the algorithm should implement non-Markov decision process in opposite to reinforcement learning \cite{KaelblingMoore}. In other words for selection of different actions at the same state of the environment an agent should have memory of previous actions. Short-term memory (STM) is extensively covered in the area of recurrent neural networks research in terms of the signal reverberation \cite{HochreiterSchmidhuber} and serial order recall \cite{BotvinickPlaut}. Theorists suggested reverberation phenomenon as significant mechanism for STM \cite{Grossberg}. Nevertheless neural principles underlying generation of the behavior as a non-linear integration of the sensory information and internal state of the network are not yet discovered. As well the question of the role STM plays in the adaptive behavior with alternative actions is still underexplored.

In this paper we present a model of an agent situated in an environment with hierarchy of goals. The behavior of the agent is controlled by recurrent neural network. We simulated evolution of agents in the model world and studied emergence of short-term memory and its role in adaptive behavior.

\section{Environment with Hierarchy of Goals}

In the current study a state of an environment in which an autonomous agent ``lives'' is represented by binary vector:

\begin{equation}
  \vec{E}\left(t\right) = \left(e_{1}\left(t\right),\cdots,e_{n^{\mathrm{env}}}\left(t\right)\right) \enspace , \enspace
  e_{i}\left(t\right) \in \left\{0,1\right\} \enspace .
\end{equation}

At any discrete time step the agent may change a single bit of this vector to the opposite. Thus, environmental structure is a $n^{\mathrm{env}}$-dimensional hypercube. Every dimension of this hypercube might be interpreted as some feature of the real environment, for instance temperature, pressure, etc. Competing goals of different complexity are prespecified in the environment as an ordered set of single bit changes of the environmental state vector:

\begin{equation}
  \vec{g}_{i} = \left(\left(n_{1},q_{1}\right),\cdots,\left(n_{k_{i}},q_{k_{i}}\right)\right) \enspace ,
\end{equation}

\noindent where $n_{j}$ is a number of target bit of the state vector, $q_{j}$ is target value of the bit, $k_{i}$ is a complexity of the goal. Goals of varying complexity are defined in the environment, together forming a branched hierarchical structure. We have generated structure of the environment by, firstly, composing hierarchy of full goals and then each goal was divided into sequence of subgoals. To determine the complexity of a particular environment we introduce occupancy based on a probabilistic approach:

\begin{equation}
  C_{\mathrm{O}} = \sum\limits_{i=1}^{N_{\mathrm{A}}}2^{k_{i}-k_{i}^{'}}\left(\frac{1}{2n^{\mathrm{env}}}\right)^{k_{i}} \enspace  ,
\end{equation}

\noindent where $N_{\mathrm{A}}$ is the number of goals in an environment, $k_{i}^{'}$ is the number of unique bits of the state vector, changed during the process of achieving a goal (single bit can be changed several times). The value, which is opposite to the occupancy, we shall call the difficulty of the environment:

\begin{equation}
  C_{\mathrm{D}} = \frac{1}{C_{\mathrm{O}}} \enspace  .
\end{equation}

Reward is associated with each goal in the environment and directly proportional to the complexity of corresponding goal. After the agent reached goal, level of accruing reward associated with this goal is reset and then linearly recovers to the original value over the time $T_{\mathrm{rec}}$.

Environment can be either deterministic or non-deterministic. In the later changes of the state vector occur not only when the agent performs actions, but also stochastically according with a fixed probability.

\section{Agent's Behavior and Evolutionary Algorithm}

Agent's behavior in the environment is controlled by artificial neural network (ANN) of arbitrary topology, which develops through the evolutionary process. ANN is divided into input, output and number of hidden layers. The network consists of McCulloch-Pitts neurons with nonnegative logistic activation function. Signal passes through the synapse only when output value of a presynaptic neuron is above a certain activation threshold (was set to 0.5 in all simulations). Thus, we introduce two possible states for each neuron: active (when output is above a threshold) and inactive (otherwise). The current state vector of the environment is directly fed to the neurons of input layer of the network. Hidden layers allow existence of recurrent connections with delayed transmission. Pair of the most active neurons of output layer encodes agent's action at the current time step. Each pair stands for a particular bit of the state vector and target value. Thereby transfers of a bit from 0 to 1 and vice versa are encoded by different pairs of the output neurons. The agent could produce ineffective actions, when attempts to change a state of a bit to one in which it already is.

Population of agents evolves in the hypercube environment described in previous section. Every individual is separately placed in the environment to evaluate fitness. For a given time the agent operates in the environment, reaching goals and accumulating reward. Value of accumulated reward is inaccessible to the agent and total reward affects its reproductive success.

To evolve agents' ANNs we used neuroevolutionary algorithm based on duplication of neurons. This algorithm is similar to well-known NEAT \cite{KennethMiikkulainen} as it makes possible evolution of neural network topology but in more natural way by duplicating neurons with attached connections. Specifically we use \textit{neuron's duplication} mutation when duplicated neuron inherits from parent the whole synaptic structure instead of \textit{add node} mutation of NEAT. Incoming connections of the descendant and parent neurons retain its previous weights, while weights of outgoing connections divided in half for both neurons. Thus, two neurons in the aggregate perform the former function, but later in evolution the descendant neuron could diverge into a separate structure. To optimize evolution by dynamically reducing dimensionality of the search space we also introduced \textit{delete connection} mutation, which performed in the same manner as \textit{add connection} mutation. Neurons that lost all their connections are being deleted from the network. We did not use any crossover algorithm, since preliminary results of comparison between two versions of evolutionary algorithm (with crossover and without) have not showed significantly difference.

The structure of the agents' networks in initial population consists of input and output layers and one additional interneuron, which will be an ancestor for all future interneurons. This initial interneuron is fully connected to input layer (by incoming connections) and to output layer, herewith there are no initial direct connections between input and output neurons.

We used the following parameters of the artificial environment in all simulations: dimensionality of the hypercube environment $n^{\mathrm{env}}$ -- 8 bits; agent lifetime duration $T_{\mathrm{life}}$ -- 250 time steps; reward's recovery time $T_{\mathrm{rec}}$ -- 30 time steps; probability of random change of the state vector's bit $P_{\mathrm{stoch}}$ -- 0.0085 (for each bit). Evolutionary algorithm was running with parameters: population $N_{\mathrm{p}}$ -- 250 agents; period of evolution $T_{\mathrm{ev}}$ -- 5000 generations; probability of the synaptic weight mutation $P_{\mathrm{m}}$ -- 0.6 (for each synapse); variance of the synaptic weight mutation $D_{\mathrm{m}}$ -- 0.08; probability of adding the synapse $P_{\mathrm{addsyn}}$ -- 0.1 (for whole network); probability of deleting the synapse $P_{\mathrm{delsyn}}$ -- 0.05 (for whole network); probability of neuron's duplication $P_{\mathrm{dup}}$ -- 0.007 (for whole network).


\section{Behavioral Evolution}

\begin{figure}[h!]
\centering
\epsfig{file=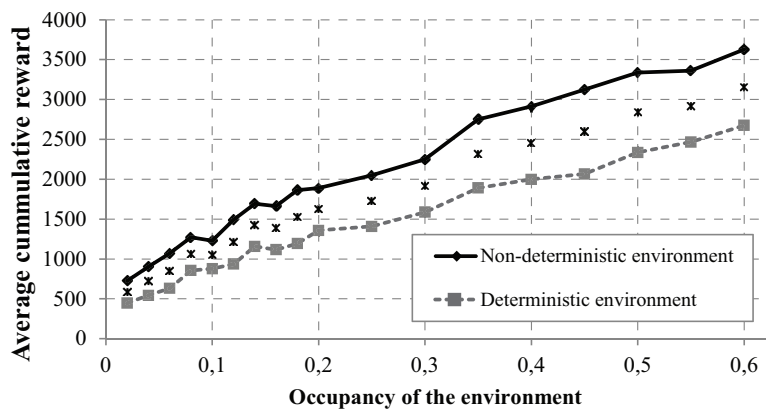, width = 10 cm}
\caption{Dependence of the average cumulative reward on the occupancy of the environment 
(each dot is averaging over 20 environments and 10 runs of the evolution in each environment, $\ast$ - t-test with $p=0.01$).}
\label{fig:statevolution}
\end{figure}

As the first step we studied how occupancy of the environment affects efficiency of evolved behavior in deterministic and stochastic environments. As expected the simulation results show the growth of an average cumulative reward with increase of the occupancy, reflecting that a task is easier when the environment is more densely populated with goals (Fig.~\ref{fig:statevolution}).

Agents evolved in non-deterministic environments behave significantly more successfully (Fig.~\ref{fig:statevolution}) in terms of average accumulated reward and in most cases have richer repertoires of behavior. Such phenomenon can be explained by the fact that during the ``life'' in an environment which could suddenly change the agent with a wider spectrum of behavioral policies should be more successful. Thus stochasticity of the environment contributes to the selection of agents with more flexible and robust behavior, allowing them to reach greater number of goals from different initial position. However, with increasing probability of random changes in the environment one can observe a sharp decline in the efficiency of the evolution (data not shown) due to the destabilization of all policies caused by excessively frequent changes in the environment.

We then analyzed evolution of behavioral sequences of the agents in the environment of intermediate goal occupancy. Typical evolved behavior usually consists of two phases: preliminary phase of converging to the main behavioral cycle and then residing on it. In a particular behavioral run both of these phases depend on the initial state of the environment, from which the agent has been started. We shall call the main behavioral cycle (sequence of actions or attained goals) a behavioral strategy.

\begin{figure}[h!]
\centering
\epsfig{file=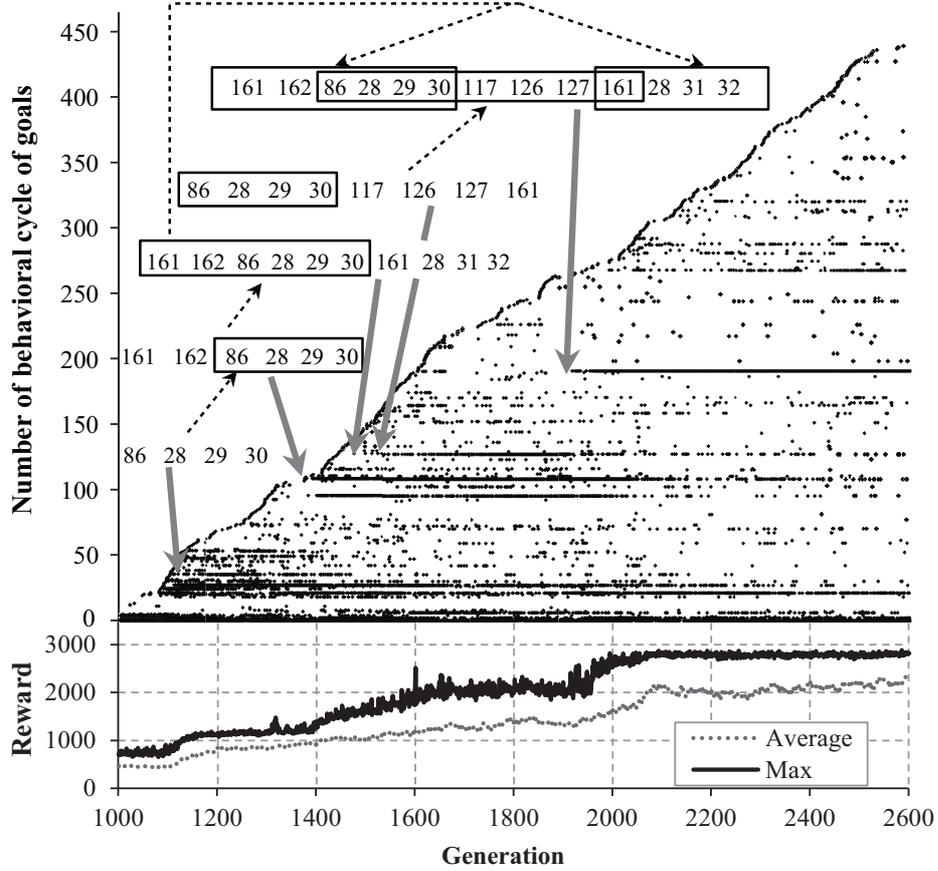, width=12.2cm}
\caption{Emergence and selection of behavioral strategies in the evolution. Sequences of goals for the most successful strategies are labeled. Each dot represents behavioral strategy; light arrows indicate moments of corresponding cycle's fixation in the evolution; black boxes indicate inclusion of old cycles into a new one.}
\label{fig:behevolution}
\end{figure}

Results of simulations demonstrate that sharp increase in the average cumulative reward among the population usually coincides with emergence of a large number of new cycles and their dynamic competition. Corresponding periods are the most interesting in the study of behavioral evolution. On the Fig.~\ref{fig:behevolution} we give an example of such period. 

One can observe competition between different behavioral strategies during a very limited evolutionary period. Herewith new strategies are appearing as an extension of previously evolved cycles. Extension can be implemented not only as simple addition of a new part to an old one, but also as complex compilation of two strategies. Eventually one of the strategies wins the competition and dominates in the behavior of the most of the population. Then evolution occurs through increase in number of states, from which it could be implemented.

\section{Alternative Behavior and Short-Term Memory}

\begin{figure}[h!]
\centering
\includegraphics{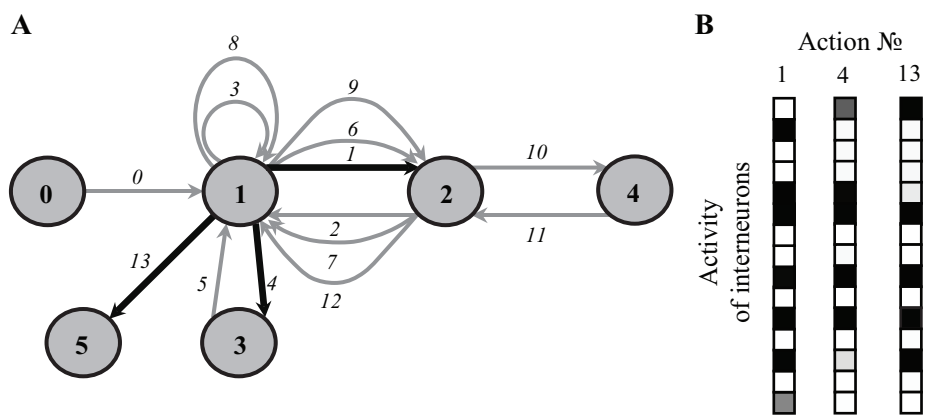}
\caption{\textbf{A)} An example of alternative behavior (states are marked as circles; transitions/actions are marked as arrows and numbered sequentially). \textbf{B)} Activity of interneurons during performance of three alternative actions (black color -– maximum activity level of corresponding neuron; white -– zero activity) }
\label{fig:alternativebehavior}
\end{figure}

The analysis of behavioral strategies emerging in the simulated evolution shows that the agents acquire ability to store short-term memory due to reverberation in the neural network by recurrent connections. The evidence for short-term memory comes from the fact that agents can implement policies based on alternative actions. Agents usually can perform different actions from one state of the environment depending on the previous history of behavior. State transition diagram for the exemplar evolved behavior with three alternative actions is shown on the Fig.~\ref{fig:alternativebehavior}. Such phenomenon would not been possible in the case of reactivity work of the neural network. Corresponding behavior is generated by a neural network consists of 30 neurons with only 15 interneurons and 611 synaptic connections.

Only a small number of neurons significantly change output and affect the decision making at states associated with alternative situations. This group of neurons is specialized on particular alternative actions. The majority of neurons remain at the same level of activity, while neurons which determine a behavior change their activity from zero to maximum level.

Based on the various behavioral policies the deepest short-term memory we found was at least 4 past states (lower bound). Such conclusion was made after considering situations in which alternative actions are performed with the same history of behavior and, thus, the lower bound can be determined as the first different action of the two behavioral sequences.

\begin{figure}[h!]
\centering
\epsfig{file=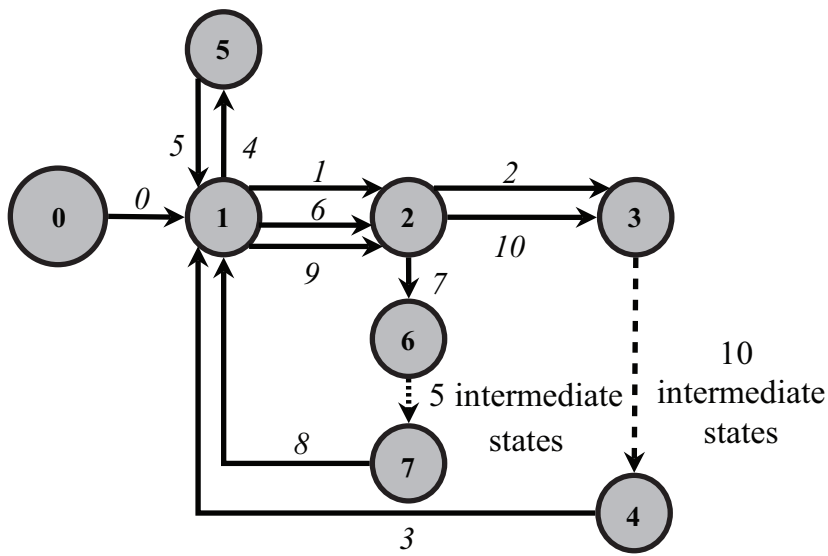, height=7.3 cm}
\caption{Behavioral strategy consisting of two cycles of actions}
\label{fig:behstrategy}
\end{figure}

The ability to use short-term memory makes possible implementation of much more complex behavior and accumulation of more reward during the agent's ``life''. Example of a behavioral strategy, based on the alternation of two cycles of actions, is shown on the Fig.~\ref{fig:behstrategy}. Since after the goal is reached it restores the reward for a fixed number of time steps, such a strategy allows goals that were reached on the first cycle to restore their value, while the agent passes the second cycle. Behavior on the Fig.~\ref{fig:behstrategy} can be implemented using short-term memory with depth of two states.

\begin{figure}[h!]
\centering
\epsfig{file=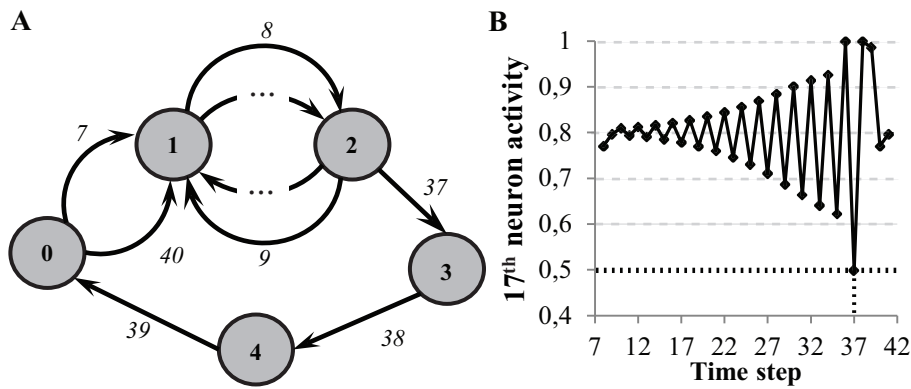}
\caption{An alternative behavior based on slow neurodynamical process. \textbf{A)} Example of alternative behavior with ``memory''. \textbf{B)} Dynamics of neuron's output responsible for realization of the first part of behavior (consecutive transitions between first and second states)}
\label{fig:slowprocess}
\end{figure}

Primitive alternative behavior in the model could also be supported by slow neurodynamical process (Fig.~\ref{fig:slowprocess}). In this case the ``memory'', required to perform alternative actions, could be up to 30 previous states. The mechanism underling this type of behavior is oscillatory dynamic of the neuron's output. While output of a neuron is above threshold, and therefore transmitted by outgoing connections, agent performs the first part of behavior. At some point output falls below the transmission threshold and the neuron no longer inhibits the other neuron, which activity is essential for the second branch of behavior.

\begin{figure}
\centering
\includegraphics{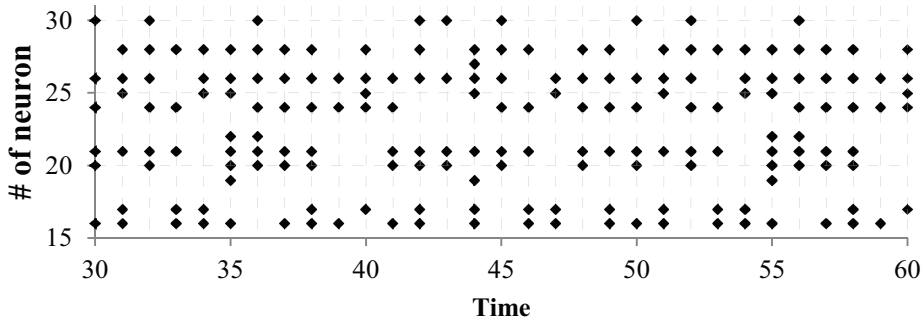}
\caption{Raster plot of a network's activity during behavior}
\label{fig:rasterplot}
\end{figure}

In the records of neural network activity (Fig.~\ref{fig:rasterplot}) during the entire period of autonomous agent's behavior one can identify both neurons, that are active while performing most of the actions, and neurons, that are active only in a very limited number of states of the environment (sometimes even in a single state).

\section{Conclusion}

In our study we have explored possibility of evolutionary emergence of short-term memory in recurrent neural networks controlling agents' behavior in multi-goal environments. On the first stage we analyzed overall dynamics of a model. As expected, performance of agents was worse when the number of goals in environment was smaller. On the other hand it turned out surprisingly that in stochastic environments agents with more flexible and stable behavioral policies evolve and able to accumulate more reward in comparison to agents in the same environment but without noise. This phenomenon is quite unusual for machine learning algorithms \cite{KaelblingMoore}, since in the case of non-deterministic environment the performance is usually worse.

Detailed investigation of evolutionary dynamics revealed that the evolution of agents consisted of two stages: rapid emergence/development of new behavioral strategies and their propagation among the population. Formation of new strategies was usually manifested as complex integration of few already existed.

On the level of individual behavior we found that evolution develops strategies, which are based on alternative actions. That happens due to acquisition of ability to operate with short-term memory and, therefore, to select actions taking into account previous history of behavior.

Evolution discovered two different neuronal mechanisms for implementation of alternative actions: first one is based on integration of sensory information and internal signal, which is reverberating through recurrent connections; and the second is based on slow neurodynamical oscillatory processes. In fact, there is no need to recruit synaptic plasticity during the agent's ``life'' to have an effective use of short-term memory. It is important to notice that emergence of the ability to operate with short-term memory in our model occurs without any artificial prerequisites in the structure of an evolutionary algorithm.

The next stage of presented study is introduction of learning during agent's ``life'', since it makes significant contribution into development of adaptive behavior. The main mechanism of learning algorithm is detection on the neuronal level of problems for the whole organism situated in environment. Possible approach to the construction of such algorithm is the use of research in the theoretical neuroscience \cite{Anokhin,Edelman}.

\subsubsection*{Acknowledgments.} This research was partially supported by Russian Foundation for Basic Research (RFBR) project 11-04-12174-ofi-m-2011.


\begin{thebibliography}{14}
\bibitem{SuttonBarto} Sutton, R.S., Barto, A.G.: Reinforcement Learning: An Introduction. MIT Press, Cambridge, MA: A Bradford Book (1998)

\bibitem{SinghBarto} Singh, S., Lewis, R., Barto, A.G.: Where Do Rewards Come From? In: Taatgen, N.A., van Rijn, H. (eds.), Proceedings of the 31st Annual Meeting of the Cognitive Science Society, pp. 2601--2606. Austin, TX: Cognitive Science Society (2009)

\bibitem{BotvinickBarto} Botvinick, M.M., Niv, Y., Barto, A.G.: Hierarchically Organized Behavior and Its Neural Foundations. A Reinforcement Learning Perspective. Cognition, vol.113, is. 3, pp. 262--280 (2009)

\bibitem{SandamirskayaSchoner} Sandamirskaya, Y., Schoner, G.: An Embodied Account of Serial Order: How Instabilities Drive Sequence Generation. Neural Networks, vol. 23, no. 10, pp. 1164--1179 (2010)

\bibitem{KomarovBurtsev} Komarov, M. A., Osipov, G. V., Burtsev, M. S.: Adaptive Functional Systems: Learning with Chaos. Chaos, vol. 20, is.4, 04511 (2010)

\bibitem{FloreanoMondana} Floreano, D., Mondana, F.: Automatic Creation of an Autonomous Agent: Genetic Evolution of a Neural-Network Driven Robot. In: Cliff, D., Husbands, P., Meyer, J.-A., Wilson, S.W. Proceedings of the 3rd International Conference on Simulation of Adaptive Behavior, pp. 421--430 (1994)

\bibitem{FloreanoMattiussi} Floreano, D., Durr, P., Mattiussi, C.: Neuroevolution from Architectures to Learning. Evolutionary Intelligence, vol. 1, no.1, pp. 47--62 (2008)

\bibitem{KaelblingMoore} Kaelbling, L.P., Littman, M.L., Moore, A.W.: Reinforcement Learning. A Survey. Journal of Artificial Intelligence Research, vol. 4, pp. 237--285 (1996)

\bibitem{HochreiterSchmidhuber} Hochreiter, S., Bengio, Y., Frasconi, P., Schmidhuber, J.: Gradient Flow in Recurrent Nets -- the Difficulty of Learning Long-Term Dependencies. A Field Guide to Dynamical Recurrent Neural Networks, IEEE Press, pp. 237--243 (2001)

\bibitem{BotvinickPlaut} Botvinick, M.M., Plaut, D.C.: Short-Term Memory for Serial Order: A Recurrent Neural Network Model. Psychological Review, vol. 113, no. 2, pp. 201--233 (2006)

\bibitem{Grossberg} Grossberg, S.: Contour Enhancement, Short-Term Memory, and Constancies in Reverberating Neural Networks. Studies in Applied Mathematics, vol. 52, no. 3, pp. 213--257 (1973)

\bibitem{KennethMiikkulainen} Kenneth, S., Miikkulainen, R.: Evolving Neural Network through Augmenting Topologies. Evolutionary Computation, vol. 10, no. 2, pp. 99--127 (2002)

\bibitem{Anokhin} Anokhin, P.K.: Biology and Neurophysiology of the Conditioned Reflex and Its Role in Adaptive Behavior. Pergamon, Oxford (1974)

\bibitem{Edelman} Edelman, G.: Neural Darwinism: The Theory of Neuronal Group Selection. NY: Basic Books (1987)

\end{thebibliography}
\end{document}